\documentclass{llncs}
\usepackage{multirow}
\usepackage{csquotes}
\usepackage{booktabs}
\usepackage{graphicx}
\usepackage{mathptmx} 
\usepackage{times} 
\usepackage{amsmath} 
\usepackage{amssymb}  

\usepackage{subfigure}
\usepackage{float}
\usepackage{booktabs}
\usepackage{mathalfa}
\usepackage[mathcal]{eucal}
\usepackage[scientific-notation=false]{siunitx}
\usepackage{color}
\usepackage[dvipsnames]{xcolor}

\usepackage[hyperfootnotes=false,hyperfigures=false]{hyperref}

\renewcommand{\thefootnote}{\fnsymbol{footnote}}

\title{Adaptive feature recombination and recalibration for semantic segmentation: application to brain tumor segmentation in MRI}

\author{S\'{e}rgio Pereira\inst{1,2} \and Victor Alves\inst{2} \and Carlos A. Silva\inst{1}}

\institute{CMEMS-UMinho Research Unit, University of Minho, Guimar\~{a}es, Portugal \\ \email{id5692@alunos.uminho.pt}; \email{csilva@dei.uminho.pt} \and Centro Algoritmi, University of Minho, Braga, Portugal}

\begin{document}

\maketitle

\renewcommand*{\thefootnote}{\arabic{footnote}}
\setcounter{footnote}{0}

\begin{abstract}

Convolutional neural networks (CNNs) have been successfully used for brain tumor segmentation, specifically, fully convolutional networks (FCNs). FCNs can segment a set of voxels at once, having a direct spatial correspondence between units in feature maps (FMs) at a given location and the corresponding classified voxels. In convolutional layers, FMs are merged to create new FMs, so, channel combination is crucial. However, not all FMs have the same relevance for a given class. Recently, in classification problems, Squeeze-and-Excitation (SE) blocks have been proposed to re-calibrate FMs as a whole, and suppress the less informative ones. However, this is not optimal in FCN due to the spatial correspondence between units and voxels. In this article, we propose feature recombination through linear expansion and compression to create more complex features for semantic segmentation. Additionally, we propose a segmentation SE (SegSE) block for feature recalibration that collects contextual information, while maintaining the spatial meaning. Finally, we evaluate the proposed methods in brain tumor segmentation, using publicly available data.

\end{abstract}

\section{Introduction}

Brain tumor segmentation plays an important role during treatment planning and follow-up evaluation. But, it is time-consuming and prone to inter- and intra-rater variability. Therefore, automatic and reliable methods are desirable. However, brain tumor segmentation is a challenging task, due to their irregular shape, appearance, and location \cite{pereira2016brain,zhao2018deep}. Recently, CNN-based approaches have achieved state of the art results \cite{pereira2016brain,zhao2018deep,kamnitsas_ensemble}. With enough data, CNNs can learn complex patterns, such as brain tumor attributes, which are, otherwise, difficult to capture by feature engineering. 

Despite being used in many applications, many CNN developments are first evaluated in image classification. In VGGNet \cite{simonyan2014very} it was shown that replacing a layer with large kernels by blocks of several layers with $3 \times 3$ kernels result in deeper and more powerful CNNs. Later, He et al. \cite{he2016deep} proposed residual learning using identity-based skip connections that allow better gradient flows and training of very deep CNNs. Other studies explored the recombination of FMs, either by compression with convolutional layers with $1 \times 1$ kernels \cite{lin2013network}, or by dividing a stack of FMs into smaller groups with grouped convolutions \cite{xie2017aggregated}. More recently, Hu et al. \cite{hu2017squeeze} proposed FM recalibration with the SE block. This block is inspired by the intuition that not all FMs are informative for all classes. Therefore, the SE block learns how to adaptively suppress the least discriminative FMs (recalibration). They showed that the simple addition of this block to state-of-the-art CNNs increased their representational power.

Semantic segmentation is one of the domains where CNNs have been pervasively used. Although one can use conventional CNNs with fully connected layers  \cite{pereira2016brain}, FCNs \cite{ronneberger2015u} are arguably one of the most important advancements regarding CNNs for semantic segmentation. In these architectures, fully-connected layers are replaced by convolutional layers, usually, with $1\times 1$ kernels. In this way, a set of voxels from an image patch can be efficiently classified in just one forward pass. In FCNs, there is a direct spatial correspondence between units in the FMs and the classified voxels. Most of the advancements in CNN design can be easily incorporated into FCN. For instance, the principles of VGGNet \cite{simonyan2014very} and residual learning \cite{he2016deep} were incorporated in \cite{pereira2016brain} and \cite{kamnitsas_ensemble}, respectively. However, although the SE block has the attractive property of re-calibrating FMs, it was conceived to weight whole FMs, which is not optimal for FCN. Since there is a spatial correspondence between units in FMs and the voxels, it is desirable to emphasize or suppress certain regions of the FMs, instead of the whole FM.

In this paper, we explore the recombination and recalibration of FMs. In recombination, instead of reducing the FMs number only, we employ linear expansion followed by compression for mixing the information. Additionally, we study how to incorporate recalibration into FCN. In SE block, global average pooling captures the whole contextual information in a FM. Instead, we argue that dilated convolution \cite{dilated} is better suited for the recalibration block in FCN. Hence, the contribution of this paper is threefold. First, we propose recombination of FMs by linear expansion and compression. Second, we explore FMs recalibration in the context of FCN. We observe that the original SE block is not optimal for FCN, and we propose a better-suited alternative. Third, we evaluate our proposal on brain tumor segmentation, using publicly available data.

\section{Methods}
\label{sec:methods}

We follow a hierarchical FCN-based brain tumor segmentation approach. Thus, we start by roughly segmenting the whole tumor with a binary FCN (WT-FCN). Using this segmentation, we define a cuboid region of interest (ROI) around the tumor with a margin of 10 extra voxels in each side. Finally, a second multi-class FCN (MC-FCN) is responsible for segmenting the multiple tumor structures inside the ROI. The proposed FCNs are inspired by an encoder-decoder architecture with long skip connections \cite{ronneberger2015u}. The input for the FCNs are image patches extracted from all the available MRI sequences. In this section, we first define the baseline FCNs; then, we present the proposed recombination and recalibration (RR) block. This block is evaluated in the more challenging multi-class segmentation problem. The WT-FCN is fixed across all experiments to isolate and make it easier to compare improvements introduced by the RR block.

\subsection{Baseline segmentation approach}
\label{sec:baseline}

\begin{figure}[t]
\centering
\includegraphics[width=1.0\textwidth, keepaspectratio]{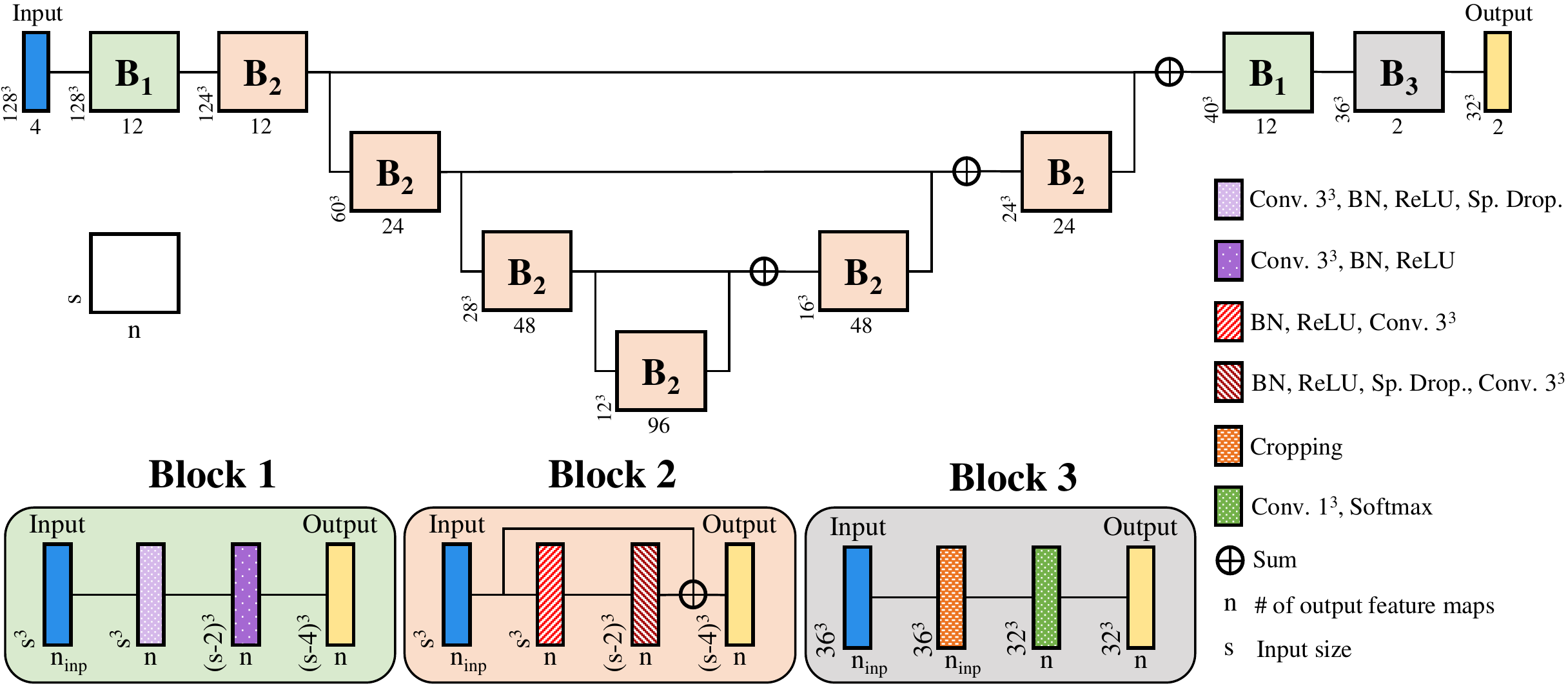}
\caption{Architecture of the WT-FCN. Downsampling is obtained by max-pooling. We use nearest neighbor upsampling to increase the FMs size, and $1\times 1\times 1$ conv. layers to adjust the number of FMs before addition. BN stands for batch normalization, and Sp. Drop. for spatial dropout.}\label{fig:grade}
\end{figure}

The architecture of the 3D WT-FCN is depicted in Fig. \ref{fig:grade}. We used both regular blocks of convolutional layers, and blocks with residual connections and pre-activation \cite{he2016deep}. This network segments 3D patches, with the three pooling layers providing a large field of view. These two characteristics contribute for reducing the number of voxels with false positive tumor detections. The baseline MC-FCN architecture can be perceived from Fig. \ref{subfig:multi_seg}, by not considering the RR block. We design the MC-FCN as a 2D network, as a proof of concept to evaluate the proposed component, which makes it computationally cheaper than the WT-FCN. Hence, 2D image patches are extracted in the axial plane. Additionally, we observed no benefits from using residual connections, or from being as deep as the WT-FCN. 

\subsection{Recombination and recalibration}
\label{sec:segse}

\begin{figure}[!htb]
\centering
\subfigure[]{\label{subfig:multi_seg}\includegraphics[width=1.0\textwidth, keepaspectratio]{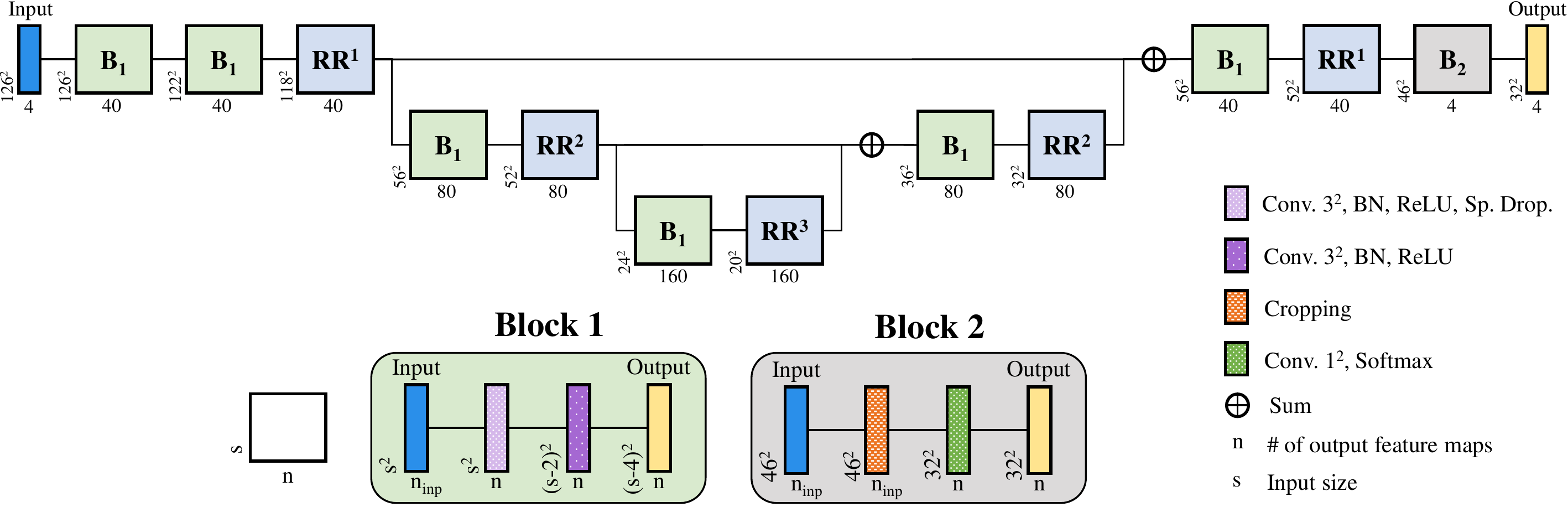}}
\subfigure[]{\label{subfig:recombination}\includegraphics[width=0.3\textwidth, keepaspectratio]{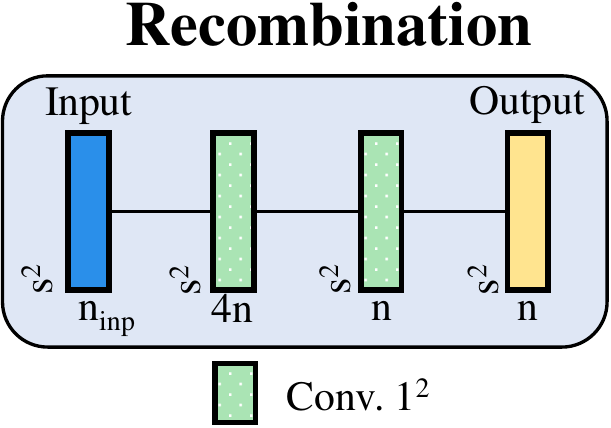}} 
\hfill
\subfigure[]{\label{subfig:recalibration}\includegraphics[width=0.9\textwidth, keepaspectratio]{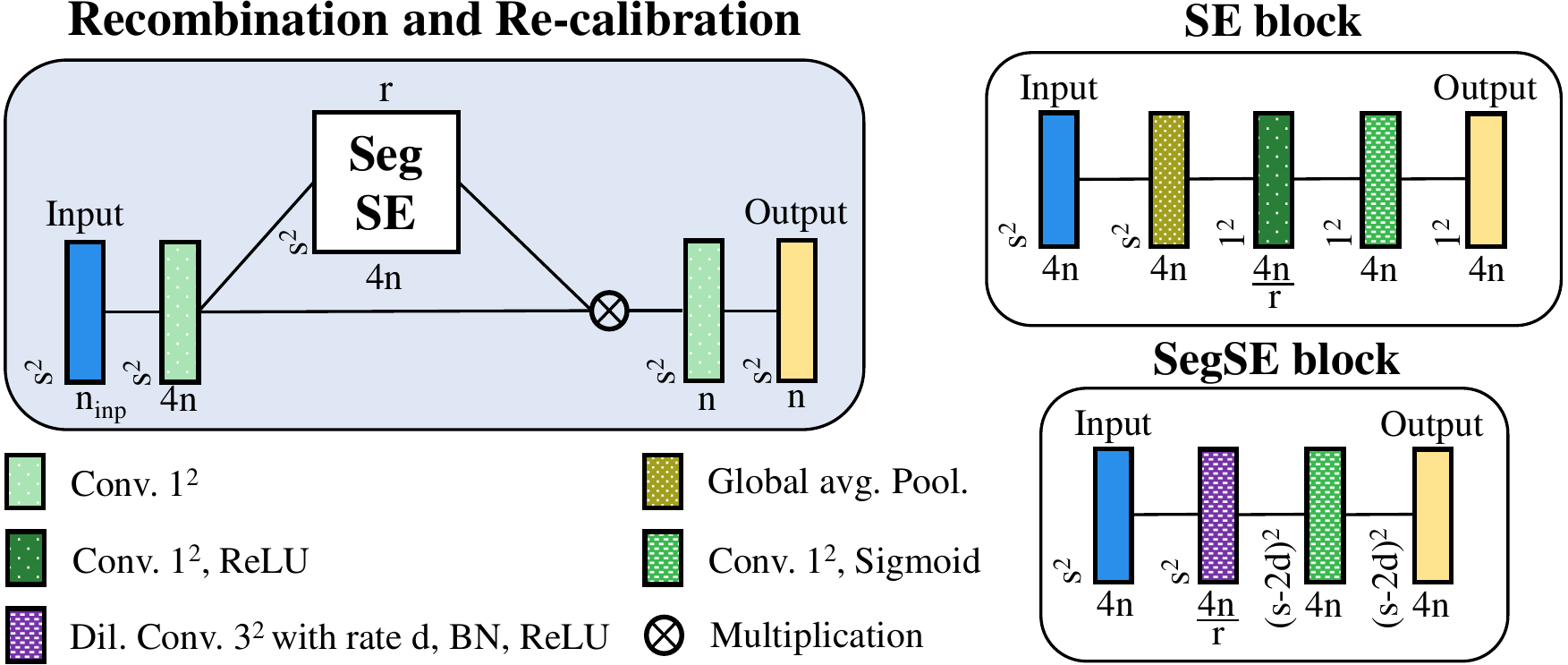}}
\caption{Architecture of the MC-FCN. a) Architecture overview with the RR block. Input sizes correspond to the RR block with SegSE. Downsampling is obtained by max-pooling. We use nearest neighbor upsampling to increase the FMs size, and $1\times 1$ conv. layers to adjust the number of FMs, before addition. b) Recombination block. c) RR block, and the SE and SegSE blocks.}\label{fig:senet}
\end{figure}

We propose recombination and recalibration of FMs as complementary operations. Recombination consists in mixing the information across FMs channels to create new combined features. In the past, convolutional layers with $1\times 1$ kernels were proposed as cross channel parametric pooling \cite{lin2013network} to decrease (compress) the number of FMs. Also, bottleneck blocks in ResNet compress the channels, process them, and expand to more channels. Instead, we propose linear recombination of FMs to increase the number of FMs (expansion), followed by compression to the original number. This operation is done by convolutional layers with $1\times 1$ kernels (Fig. \ref{subfig:recombination}). Experimentally, we found an expansion factor of 4 to work well. Linear recombination of units in a given spatial location of the FMs results from the weighted sum of the units in the same location of all FMs. Hence, expansion combines features into a higher dimension, while compression learns how to compress features and suppress the least discriminative ones. 

We propose the RR block (Fig. \ref{subfig:recalibration}) that combines both FM recombination and recalibration. Feature map recalibration with the SE block, as proposed in \cite{hu2017squeeze}, is shown in Fig. \ref{subfig:recalibration} -- SE block. First, global average pooling summarizes each FM into its average value to capture contextual information. Then, two fully connected (FC) layers\footnote{Equivalently implemented as convolutional layers with $1\times 1$ kernels.} capture cross-channels relations. The first one is a compression layer with a factor of $r$, followed by ReLU activation. The second FC layer restores the original dimension, and is followed by the sigmoid activation function. Finally, this vector is channel-wise multiplied with the input FMs, i.e., each FM is multiplied by a corresponding scalar value, resulting in each FM being scaled as a whole. Ideally, less discriminative FMs are suppressed. This approach was shown to improve learning in image classification. In this problem, a FM may have a strong response for a given class or subset of classes. However, in semantic segmentation with FCN, a patch of voxels is segmented at once. So, there is a correspondence between segmented pixels and units in FMs. In this scenario, some regions of the FMs may have strong activations that are relevant for the structure that is being segmented in that spatial location. Hence, the SE block may be not optimal for semantic segmentation, since it collapses the whole FM into a single value, regardless of the regions. Thus, the proposed RR block includes a segmentation adapted SegSE block. A straightforward approach for adapting the SE block for semantic segmentation is by simply removing the global average pooling layer. In this way, the spatial correspondence among units and voxels is maintained. However, this is also not optimal, since contextual information is important to evaluate the spatial importance of a given feature. In preliminary experiments, this approach resulted in worse performance. Therefore, we propose our SegSE block (Fig. \ref{subfig:recalibration}) that uses a convolutional layer with $3\times 3$ kernels with dilation $d$ for context aggregation. Simultaneously, this layer is responsible for the compression stage. Experimentally, we found that the best dilation rates depend on the resolution of the FMs. This is due to the fact that deeper layers already have a larger field of view. Hence, we set the rate in $\{RR^1, RR^2, RR^3\}$ (c.f. Fig. \ref{subfig:multi_seg}) to $\{3, 2, 1\}$. In preliminary experiments, we evaluated using spatial average pooling followed by convolutional and transposed convolutional layers, but, we obtained worse performance. The reason is probably due to the checkerboard artifacts that appear with this combination of layers, but not with dilated convolution.

\section{Experimental Setup}
\label{sec:exp_set}

We evaluate the proposed blocks in the Brain Tumor Segmentation Challenge (BRATS) 2017 and 2013 databases \cite{bakas2017advancing,menze2015multimodal}. BRATS 2017 has two publicly available datasets: Training (285 subjects) and Leaderboard (46 subjects). In BRATS 2013 we use Training (30 subjects) and Challenge (10 subjects). For each subject, there are four MRI sequences available: T1, post-contrast T1 (T1c), T2, and FLAIR. All images are already interpolated to $1 mm$ isotropic resolution, skull stripped, and aligned. Only the Training sets contain manual segmentations. In BRATS 2017 it distinguishes three tumor regions: edema, necrotic/non-enhancing tumor core, and enhancing tumor. In BRATS 2013 the manual segmentation have necrosis and non-enhancing tumor separately, although we fuse these labels to be similar to BRATS 2017. Evaluation is performed for the whole tumor (all regions combined), tumor core (all, excluding edema), and enhancing tumor. Since annotations are not publicly available for 2017 Learderboard and 2013 Challenge, the evaluation is computed by the CBICA IPP and SMIR online platforms\footnote{https://ipp.cbica.upenn.edu/ and https://www.smir.ch/BRATS/Start2013}. The development of the RR block was conducted in the larger BRATS 2017 Training set, which was randomly divided into training (60\%), validation (20\%), and test (20\%)\footnote{Subjects id in each set are available online: \url{https://github.com/sergiormpereira/rr_segse}.}. However, networks tested in BRATS 2013 Challenge were trained in the 2013 Training set.

Image pre-processing included bias field correction, and standardization of the intensity histograms of each MRI sequence, as in \cite{pereira2016brain}. During training, we use the crossentropy loss, the Adam optimizer with learning rate of \num{5e-5}, weight decay of \num{1e-6}, and spatial dropout probability of 0.05. Since we used convolution without padding, during multiplication or sum of FMs with different sizes, we cropped the center part of the biggest one. The compression factor $r$ in the SE and SegSE blocks was set to 10. Data augmentation included sagittal flipping and random rotations of $\ang{90}$. For training the binary whole tumor FCN, all tumor regions in manual segmentations were fused into a single label. All the hyperparameters were found using the validation set, before evaluation in the test set. The FCNs were implemented using Keras and Theano.

Metrics provided by the online evaluation platforms differ in BRATS 2017 and 2013. Hence, in BRATS 2017 we use Dice and the 95\textsuperscript{th} percentile of the Hausdorff Distance (HD$_{95}$). In BRATS 2013, the online platform computes Dice, Sensitivity, and Positive Predictive Value (PPV).

\section{Results and Discussion}
\label{sec:exp_res}

\begin{figure}[t]
\centering
\includegraphics[width=0.9\textwidth, keepaspectratio]{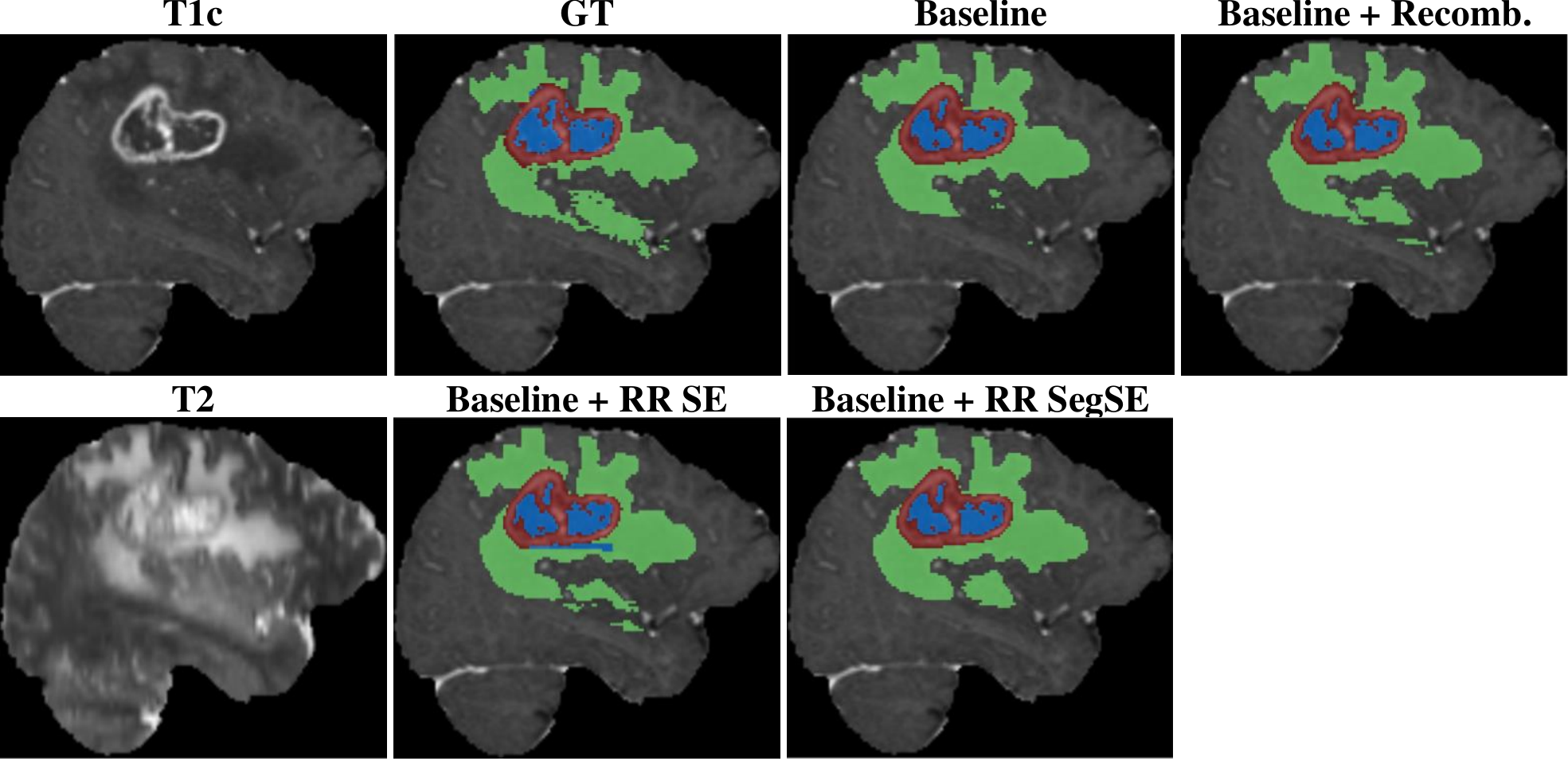}
\caption{Examples of the segmentation obtained with each of the evaluated RR blocks. The colors in segmentations mean: green -- edema, blue - tumor core, and red -- enhancing tumor.}\label{fig:segs}
\end{figure}

We evaluate the effect of recombination and recalibration of FMs using the SegSE block in the test set (20\% of BRATS 2017 Training), and compare it with the baseline and RR with SE block. Quantitative results are presented in Table \ref{tab:res_test}, and segmentation examples in Fig. \ref{fig:segs}. When we include the recombination by expansion followed by compression stage to the baseline FCN, we observe that Dice improves in all tumor regions. Although this improvement is negligible in the enhancing region, it is substantial in the whole tumor and in the core. In fact, it achieves the highest Dice for tumor core of all the blocks. However, the HD$_{95}$ is higher in all classes, when compared with the baseline.

\begin{table}[b]
\centering

\caption{Results (average) obtained in the test set (20\% of BRATS 2017 Training). We evaluate recombination (Recomb.) of FMs, and RR using both SE and SegSE blocks. Bold results show the best score for each tumor region.}

\setlength{\tabcolsep}{0.5em}
\resizebox{0.65\textwidth}{!}{\begin{tabular}{lccccccccc}
\specialrule{.2em}{.1em}{.1em}
 &       & \textbf{Dice}  &       &       & \textbf{HD$_{95}$}  &  \\ \cmidrule(l){2-4} \cmidrule(l){5-7}
    \textbf{Method}     & \textbf{Whole} & \textbf{Core}  & \textbf{Enh.} & \textbf{Whole} & \textbf{Core}  & \textbf{Enh.} \\ \midrule
    Baseline & 0.857 & 0.739 & 0.682 & 8.645 & 10.761 & 6.672 \\
    \textbf{Baseline + Recomb.} & 0.865 & \textbf{0.769} & 0.687 & 9.720 & 11.453 & 7.790 \\ \midrule
    Baseline + RR SE & 0.859 & 0.756 & 0.672 & 8.939 & 13.306 & 7.319 \\
    \textbf{Baseline + RR SegSE} & \textbf{0.866} & 0.766 & \textbf{0.698} & \textbf{8.475} & \textbf{10.513} & \textbf{6.131} \\ \bottomrule
\end{tabular}}
\label{tab:res_test}
\end{table}

In Table \ref{tab:res_test} we can find results for the recalibration stage, when joined with recombination to form the RR block. We observe that the SE block, proposed in \cite{hu2017squeeze}, leads to worse Dice, when compared to the baseline with recombination layers. Actually, the Dice of the enhancing tumor is even lower than the baseline. This may be due to enhancing region usually being a smaller part of the whole tumor volume. Additionally, finer details are needed to define this region, hence its contribution to a whole feature map response may be less strong than the other tumor regions and end up being suppressed by the SE block. Therefore, we conclude that the SE block, acting as whole FM recalibration is not optimal for segmentation. Finally, it is possible to observe from Table \ref{tab:res_test} that the RR block with the proposed SegSE recalibration achieves the best scores, both in Dice (excepting core by a small margin) and HD$_{95}$. We note that the SegSE block is the only approach that substantially improves the Dice of enhancing tumor over the baseline. Moreover, the HD$_{95}$ suggests that besides achieving good overlap scores, it also obtains the best contour definition. The SegSE stage comes at the cost of more parameters. In order to evaluate if its performance is due to these extra capacity, we proportionally increased the width of the baseline, such that its parameters number becomes similar to the network with RR + SegSE. The results obtained with this larger network in terms of Dice/HD$_{95}$ were 0.852/9.049, 0.751/10.647, and 0.678/7.065 for the complete, core, and enhancing regions, respectively. So, the RR SegSE block improvements are due to better learning, and not directly to the higher capacity.

We compare with the state of the art in BRATS 2017 Leaderboard and BRATS 2013 Challenge in Tables \ref{tab:res_leaderboard} and \ref{tab:res_2013}, respectively. In BRATS 2017, most of the top performing methods are ensembles of FCN. In principle, taking a CNN or FCN and building an ensemble will certainly lead to better results. Since we are evaluating the effect of the SegSE block, we need to assess it in a single model. So, we compare our results with other single CNN approaches, such as Islam \cite{islam} and Jesson \cite{jesson}, for the sake of fairness. Nevertheless, we present results obtained by the multi-model and multi-training settings ensemble proposed by Kamnitsas et al. \cite{kamnitsas_ensemble}, the winner of BRATS 2017 Challenge. In the single network approaches, Islam \cite{islam} employed a hypercolumns-inspired CNN. Jesson \cite{jesson} used a FCN with multiple prediction layers and loss functions in different scales. Additionally, the authors employed a learning curriculum to deal with class imbalance. From Table \ref{tab:res_leaderboard}, we observe that the baseline achieves competitive results, when compared with the single-model approaches. The Dice is comparable with Islam, while the HD$_{95}$ scores are smaller. Regarding the RR SegSE block, we confirm that it improves the baseline performance. Indeed, the results are competitive with Jesson, with better Dice for core and enhancing regions, and HD$_{95}$ in the enhancing region. In BRATS 2013 Challenge (Table \ref{tab:res_2013}), the proposed FCN with the RR SegSE block improves over the baseline, again. The other compared methods are all recent and top performing CNN-based approaches. Pereira \cite{pereira2016brain} uses a plain CNN with fully-connected layers. Shen \cite{shen} uses a FCN enhanced by input symmetry maps and a boundary-aware loss function. Zhao \cite{zhao2018deep} also proposes a FCN followed by a conditional random field trained as recurrent neural network and a sophisticated post-processing stage. We note that the proposed method achieves the highest Dice and Sensitivity scores. In fact, the baseline with the RR SegSE block is ranked 1\textsuperscript{st} by the online evaluation platform.

\begin{table}[!htb]
\centering

\caption{Results (average) obtained in BRATS 2017 Leaderboard set. Bold results show the best score for each tumor region. Underlined scores are the best among single-model approaches (excluding Kamnitsas).}

\setlength{\tabcolsep}{0.5em}
\resizebox{0.75\textwidth}{!}{\begin{tabular}{lccccccccc}
\specialrule{.2em}{.1em}{.1em}
          &       & \textbf{Dice} &       &       & \textbf{HD$_{95}$} &  \\ \cmidrule(l){2-4} \cmidrule(l){5-7}
    \textbf{Method} & \textbf{Whole} & \textbf{Core}  & \textbf{Enh.} & \textbf{Whole} & \textbf{Core}  & \textbf{Enh.} \\ \midrule
	Islam \cite{islam} & 0.876 & 0.761 & 0.689 & 9.820 & 12.361 & 12.938 \\    
    Jesson \cite{jesson} & \underline{0.899} & 0.751 & 0.713 & \underline{\textbf{4.160}} & \underline{8.650} & 6.980 \\
    Kamnitsas \cite{kamnitsas_ensemble} & \textbf{0.901} & \textbf{0.797} & \textbf{0.738} & 4.230 & \textbf{6.560} & \textbf{4.500} \\ \midrule
    Baseline & 0.878 & 0.760 & 0.692 & 6.597 & 11.915 & \underline{5.978} \\
    \textbf{Baseline + RR SegSE} & 0.884 & \underline{0.771} & \underline{0.719} & 6.202 & 10.215 & 6.702 \\ \bottomrule
\end{tabular}}
\label{tab:res_leaderboard}
\end{table}

\begin{table}[!htb]
\centering

\caption{Results (average) obtained in BRATS 2013 Challenge set. Bold results show the best score for each tumor region.}

\setlength{\tabcolsep}{0.5em}
\resizebox{0.9\textwidth}{!}{\begin{tabular}{lccccccccc}
\specialrule{.2em}{.1em}{.1em}
 &  & \textbf{Dice} &  &  & \textbf{PPV} &  &  & \textbf{Sensitivity} &  \\ \cmidrule(l){2-4} \cmidrule(l){5-7} \cmidrule(l){8-10}
\textbf{Method} & \textbf{Whole} & \textbf{Core} & \textbf{Enh.} & \textbf{Whole} & \textbf{Core} & \textbf{Enh.} & \textbf{Whole} & \textbf{Core} & \textbf{Enh.} \\ \midrule
Pereira \cite{pereira2016brain} & 0.88 & 0.83 & 0.77 & 0.88 & \textbf{0.87} & 0.74 & 0.89 & 0.83 & 0.81 \\ 
Shen \cite{shen} & 0.88 & 0.83 & 0.76 & 0.87 & \textbf{0.87} & 0.73 & 0.9 & 0.81 & 0.81 \\ 
Zhao \cite{zhao2018deep} & 0.88 & \textbf{0.84} & 0.77 & \textbf{0.9} & \textbf{0.87} & \textbf{0.76} & 0.86 & 0.82 & 0.8 \\ \midrule
Baseline & 0.87 & 0.83 & 0.77 & 0.81 & 0.81 & 0.71 & \textbf{0.94} & 0.88 & 0.87 \\ 
\textbf{Baseline + RR SegSE} & \textbf{0.89} & \textbf{0.84} & \textbf{0.78} & 0.86 & 0.83 & 0.71 & 0.93 & \textbf{0.89} & \textbf{0.88} \\ \bottomrule
\end{tabular}}
\label{tab:res_2013}
\end{table}

\section{Conclusion}

Recalibration of FMs has the power to adaptively emphasize discriminative FMs and suppress the uninformative ones. However, this is not optimal in the context of FCN for segmentation. In this work, we propose recombination and recalibration of FMs for semantic segmentation. The former employs linear expansion followed by compression of FMs for mixing features, while the later adaptively recalibrates regions of the FMs. We show that both recombination and recalibration improve over a competitive baseline. Although we opted for a simple U-net inspired network, the proposed block can be used in other more complex FCN. Still, our FCN with the RR SegSE block achieves competitive results in BRATS 2017 Leaderboard, when compared with other single-model approaches, and superior results in BRATS 2013 Challenge.

\subsubsection{Acknowledgments}

S\'ergio Pereira was supported by a scholarship from the Funda\c{c}\~ao para a Ci\^encia e Tecnologia (FCT), Portugal (scholarship number PD/BD/105803/2014). This work has been supported by COMPETE: POCI-01-0145-FEDER-007043 and FCT – Funda\c{c}\~ao para a Ci\^encia e Tecnologia within the Project Scope: UID/CEC/00319/2013.

\bibliographystyle{splncs03} 
\bibliography{references}

\begin{thebibliography}{10}
\providecommand{\url}[1]{\texttt{#1}}
\providecommand{\urlprefix}{URL }

\bibitem{bakas2017advancing}
Bakas, S., et~al.: Advancing the cancer genome atlas glioma mri collections
  with expert segmentation labels and radiomic features. Scientific data  4
  (2017)

\bibitem{he2016deep}
He, K., Zhang, X., Ren, S., Sun, J.: Identity mappings in deep residual
  networks. In: European Conference on Computer Vision. pp. 630--645. Springer
  (2016)

\bibitem{hu2017squeeze}
Hu, J., et~al.: Squeeze-and-excitation networks. arXiv:1709.01507  (2017)

\bibitem{islam}
Islam, M., Ren, H.: Multi-modal pixelnet for brain tumor segmentation. In:
  Brainlesion: Glioma, Multiple Sclerosis, Stroke and Traumatic Brain Injuries.
  pp. 298--308 (2018)

\bibitem{jesson}
Jesson, A., Arbel, T.: Brain tumor segmentation using a 3d fcn with multi-scale
  loss. In: Brainlesion: Glioma, Multiple Sclerosis, Stroke and Traumatic Brain
  Injuries (2018)

\bibitem{kamnitsas_ensemble}
Kamnitsas, K., et~al.: Ensembles of multiple models and architectures for
  robust brain tumour segmentation. In: International MICCAI Brainlesion
  Workshop (2018)

\bibitem{lin2013network}
Lin, M., et~al.: Network in network. arXiv preprint arXiv:1312.4400  (2013)

\bibitem{menze2015multimodal}
Menze, B.H., et~al.: The multimodal brain tumor image segmentation benchmark
  (brats). IEEE T Med Imaging  34(10) (2015)

\bibitem{pereira2016brain}
Pereira, S., et~al.: Brain tumor segmentation using convolutional neural
  networks in mri images. IEEE T. Med. Imaging  35(5),  1240--1251 (2016)

\bibitem{ronneberger2015u}
Ronneberger, O., et~al.: U-net: Convolutional networks for biomedical image
  segmentation. In: MICCAI. pp. 234--241. Springer (2015)

\bibitem{shen}
Shen, H., et~al.: Boundary-aware fully convolutional network for brain tumor
  segmentation. In: MICCAI. pp. 433--441 (2017)

\bibitem{simonyan2014very}
Simonyan, K., Zisserman, A.: Very deep convolutional networks for large-scale
  image recognition. arXiv:1409.1556  (2014)

\bibitem{xie2017aggregated}
Xie, S., et~al.: Aggregated residual transformations for deep neural networks.
  In: CVPR. pp. 5987--5995 (2017)

\bibitem{dilated}
Yu, F., Koltun, V.: Multi-scale context aggregation by dilated convolutions.
  In: ICLR (2016)

\bibitem{zhao2018deep}
Zhao, X.o.: A deep learning model integrating fcnns and crfs for brain tumor
  segmentation. Med. Image Anal.  43,  98--111 (2018)

\end{thebibliography}

\end{document}